# Fully Convolutional Networks for Automatically Generating Image Masks to Train Mask R-CNN

Hao Wu[1] [0000-0003-4537-2898] , Jan Paul Siebert [2] Xiangrong Xu[1*]

[1] Anhui University of Technology, Anhui, China.
[2] University of Glasgow, Glasgow, United Kingdom

*corresponding xuxr88@qq.com

**Abstract:** This paper proposes a novel automatically generating image masks method for the state-of-the-art Mask R-CNN deep learning method. The Mask R-CNN method achieves the best results in object detection until now, however, it is very time consuming and laborious to get the object Masks for training, the proposed method is composed by a two-stage design, to automatically generating image masks, the first stage implements a fully convolutional networks (FCN) based segmentation network, the second stage network, a Mask R-CNN based object detection network, which is trained on the object image masks from FCN output, the original input image, and additional label information. Through experimentation, our proposed method can obtain the image masks automatically to train Mask R-CNN, and it can achieve very high classification accuracy with an over 90% mean of average precision (mAP) for segmentation.

## 1. Introduction

Object detection represents a problem of fundamental significance in computer vision and robotics research, which combined segmentation and recognition together, its accuracy and real-time is most important ability, in recent years, object detection is used in wide range of applications in areas such as self-driving, artificial intelligence, face recognition, however, various factors will interferences the detection process, such as angle, occlusion, uneven illumination and other factors.

Krizhevsky [18] proposed a large, deep convolutional neural network named AlexNet, which achieved the state of art results in ImageNet classification, then many types of convolutional neural network are used for object recognition since then. The R-CNN serials network achieve a well performance in recognition, especially, the Mask R-CNN method gets the state of art results in object recognition, however, it is very time consuming and laborious to get the object Masks for training, so we proposed a method to automatically generating image masks to train Mask R-CNN based on fully convolutional networks. Since this method can automatically generate an image Mask, which meets the requirements of the Mask used for training Mask R-CNN, it is suitable for the field of robotic vision in object recognition, as it can now be trained to recognize new objects quickly and simply.

The network structure for object detection can be divided into one stage framework and two stage framework[1]. Two stage includes R-CNN series framework, one stage package including YOLO, SSD and other series of frameworks [2-5]. The difference between one stage and two stage is that, the former is to generate multiple candidate frames in the network to locate the object directly after inputting the image. The latter is to extract features in the convolutional neural network after inputting the image, then predict object classification and position.

The one-step detection method represented by YOLO is notable high detection speed and good real-time detection. The YOLO algorithm proposed by Redmon[6] continues the core idea of the GoogleNet model [7], treat the object detection as a regression problem, directly regress the object bounding box and the classification on the grid. Its structure is simple, the algorithm input an image, scale it to a uniform size, and divide into the S×S grids (S=7). If the center of the object to be detected falls into the grid, it is detected and classified. Compared YOLO and several common detection frameworks (DPM, R-CNN, OverFeat), YOLO has three advantages, first, one step detection. The previous optimized inspection pipeline is completely abandoned by the YOLO and YOLO perform feature extraction, bounding box prediction, and using a single convolutional neural network Non-maximum suppression simultaneously. Second, reduce the candidate box. Network proposal unit sets the space limit, and there are only 98 candidate boxes for each image extraction and improve detection efficiency. Third, YOLO is generalized and does not require multiple detectors, can detect various objects simultaneously.

The R-CNN framework is one of the famous method in two stage framework, the R-CNN framework is proposed by Girshick et al[8] in 2013, a series based on this framework is constantly being proposed[9-12], The R-CNN method also is the first deep learning-based object detection, the R-CNN algorithm steps include: (1) using selectivity Search (SS) algorithm to generate multiple candidate window in one image (generally bout 2000); (2) The feature map of each candidate window is scaled to the same size, use the Convolutional Neural Network (CNN) to extract features; (3) extract the special features collected into the SVM classifier to determine whether it belongs to a specific class; (4) make Use the regression to further adjust the candidate window position.

Based on R-CNN, Girshick et al. proposed Fast R-CNN [13], introduce object area pooling (ROI). This is actually a single pyramid pool layer solves the problem of repeated calculation of candidate box. Fast R-CNN can select proposals appropriately based on input image, In the training process, multi-task training is used instead of staged training, and the classification and regression tasks are at the same time, expand the data set (VOC10 and VOC12) to provide more training data with high training accuracy, using Softmax classifier instead of SVM classifier, the broken SVD compresses the fully connected layer into two fully connected layers that have no linear relationship.

Ross B. Girshick proposed the Faster R-CNN in 2016[14], the main part of Faster R-CNN method is the sliding window, and the core is the regional proposal. The Region Proposal Network (RPN) [15] is a full convolutional neural network. Unlike Fast R-CNN, Faster R-CNN is delivered through a four-step training method for optimization, learning the sharing between RPN and region-based object detection R-CNN, the conv layer feature eventually forms a conv layer shared by two networks. During the training process, each task cooperates with each other and shares parameters, so Faster R-CNN reduced inspection time.

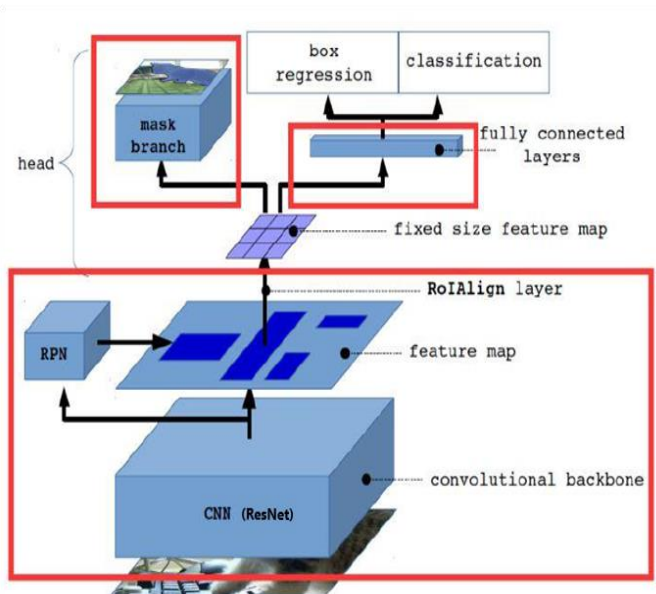

Figure 1.The MaskR-CNN framework [21]

Mask R-CNN model is extended from Faster RCNN. On each RoI in the Faster RCNN model, a mask prediction branch running in parallel with the detection branch is added[16], as shown in Figure 1. The mask branch is a small FCN network. It is applied to each RoI to predict a split mask in a pixel-to-pixel manner. Mask R-CNN is easy to implement, has little extra overhead, and has great flexibility. It is a general model for segmentation. RoI pooling used in Faster RCNN is a rough pooling method for object detection tasks, which will cause a certain degree of misalignment. To overcome this, Mask R-CNN proposes RoIAlign, which is used to preserve accurate space position, RoIAlign can increase the accuracy of the mask by 10%~50%.

The proposed method for automatically generating Mask consists of two stages to automatically generate an image mask. The first stage is to generate a mask based on a fully convolutional networks (FCN) segmentation network, and the second stage is based on a Mask R-CNN target detection network. That is, use the output Mask of the FCN of the first stage, the original input image, and the attached label information for training.

The organization of this paper is as follows: the next section will discuss in detail the proposed method which automatically generating image masks to train Mask R-CNN, and then with experimental results as well analysis.

## 2. The Specific Architecture of Proposed Method

As shown in Figure 2, the proposed method is composed by a two-stage design, the first stage implements a Fully convolutional networks (FCN) based segmentation network[17] that outputs a pixel level segmentation result. Training this FCN network with MIT scene parsing training dataset, then output the object Mask of the input image, however, there are still some interfering subregion in the output object Mask image, after blob detection with Opencv IP (image processing), the object Mask image is obtained satisfied. The second stage, where Mask R-CNN based object detection is performed, includes an additional label information that is built on the object Mask image and the original input image, and uses them as training, output the segmentation results of object as well as label information and scores. The first-stage network is referred to as the segmentation network, while the second stage network, as the object detection network.

### 2.1 FCN based segmentation network

The FCN Segmantic Segmentation algorithm was proposed by Long et al. in 2015. It is the first real implementation of full-convolutional neural networks with end-to-end training on pixel-level prediction tasks. FCN algorithm replaces the original fully connected layer in CNN network by the 1*1 convolutional layer to implement a full convolutional network. With this operation, FCN can produce a prediction result for each pixel on images, and finally outputs a prediction graph related to the input image size. In the prediction stage, the author mentions a learnable bilinear upsampling method is presented, which is proved better than the upsampling method with fixed parameters by experiments. To make the segmentation result more refined, FCN uses a layered structure that takes features from multiple layers after sampling to the same size, then they are merged by summation. For the addition of low layer information, the semantic and spatial precision of the output results are improved effectively. Another advantage of FCN is that its input image can be any size, the traditional CNN due to the existence of the fully connected layer, its input image size only can be fixed.

### 2.2 Mask R-CNN based object detection network

As mentioned before, Mask R-CNN is an improved network based on Faster RCNN Network model. adding a parallel Mask segmentation output branch, Mask R-CNN's network structure can be divided into four parts: the base network part, the RPN layer, the RoiAlign layer, and the output layer at the end of the network.

The original base network used by Mask R-CNN is ResNet, the feature map outputs corresponding to five different scales are respectively used to establish the feature pyramid structure. The five different scale feature maps extracted by the basic network layer are inputted into the RPN layer and a recommendation window is generated. Then perform a RoiAlign operation on the candidate window generated by the RPN layer.

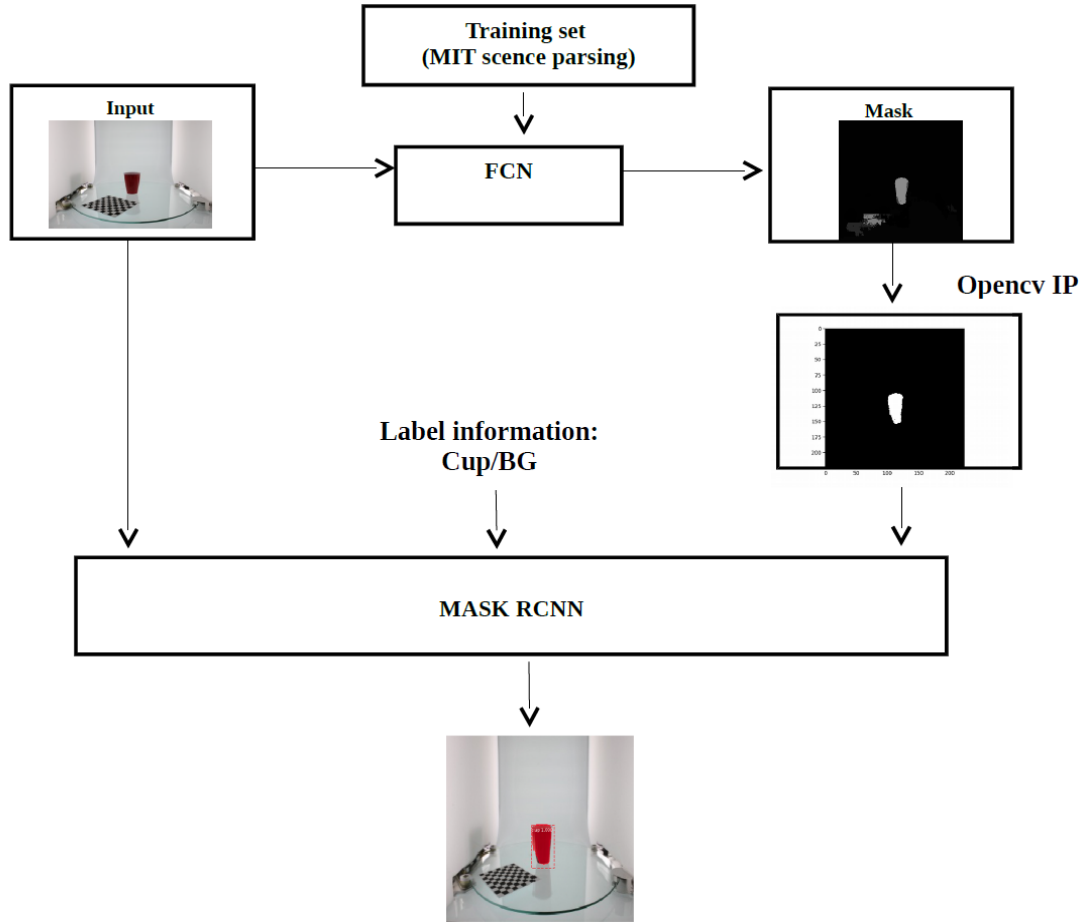

Figure 2. The overall architecture of the proposed method

The generated candidate window is caused to produce a fixed size feature map. Connect the resulting regions of interest and output the results in parallel, including the fully connected prediction branch, fully connected prediction rectangle offset branch and full convolution prediction pixel segmentation branch.

The specific steps of RoiAlign are: (1)Each candidate area is represented by a floating point number at the boundary; (2) the candidate area is divided into thousands of units, the boundary of each element is also represented by a floating point number; (3) using bilinear interpolation to obtain a fixed four coordinates of the positions in each unit, then the maximum pooling, mapping the region of interest back to the original image.

## 3. Experimental and discussion

This section mainly discuss the proposed method for segmentation and object recognition, the segmentation network is firstly trained and evaluated on the MIT scene parsing dataset[19],this dataset contains over 20K images with semantic categories, such as sky, road, person, and bed. The size of the images in the dataset is 1920 ×1080.

### 3.1FCN for segmentation

**Training**

Use VGG16 trained model to do initialization, and fine-tune all layers, add upsampling at the end, the parameter learning is trained by the back propagation principle of CNN. Fine-tune is to use the training samples to train the new model based on the trained model. It is equivalent to using the trained model to initialize the network parameters, and extract the shallow features through these parameters, and according to the self. The characteristics of the data fine-tune these parameters, avoiding the situation that the model cannot converge due to the parameter initialization problem, thus improving the training efficiency.

**Experimental platform parameters**

This FCN methods' training uses Tensorflow's python development interface under Ubuntu operating system, using 1 GPU for training, specific hardware and software configuration information is GPU NVIDIA GTX1080TI, Python 3.5,TensorFlow-gpu 1.0.

AS shown in Fig3, our proposed FCN based segmentation method have a well performance in cup, but the segmentation result still has some sub area, as shown in the bottom left part of the chekerboard in Fig 3 (b), so the image processing method with opencv library is used to extract the final object mask, which is shown in Fig 3 (c).

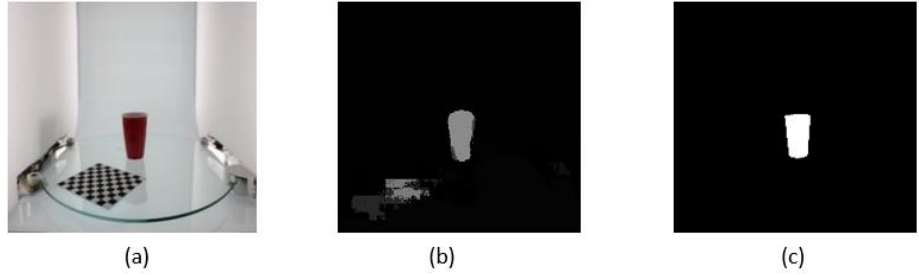

Figure 3. Segmentation results with FCN

### 3.2 Mask R-CNN for object detection
**Training**

Firstly, the Mask R-CNN model is trained and initialized on the Microsoft COCO dataset, the Microsoft COCO dataset contains 330k images with 80 categories, training on relatively large COCO dataset make sure the Mask R-CNN method is initialized to localize common objects, then the Mask R-CNN detection method is fine-tuned on the MIT scene parsing dataset, since the object class of COCO dataset is 80,the output of this layer is resized for two output class namely detection object and background. The object detection system is also evaluated on a single NVIDIA GTX 1080TI GPU, Python 3.6,TensorFlow-gpu 1.9,Keras 2.1.
Detection results with Mask R-CNN

As shown in Fig 4, the proposed detection method have a good performance in position and segmentation of cup with different angle, the mean of average precision (mAP) is used as a metric[20], the proposed method achieved mAP of 0.9985 for cup with different rotation angle.

**Error Analysis and Improvement**

The cup object can be segmented with FCN and then detection with Mask R-CNN since the COCO dataset already includes the cup objects, but the FCN cannot segment well for objects which is not exist in COCO dataset, as shown in Fig 5, the 3M high tack spray adhesive bottle object cannot be segmented totally. A good object segmentation results is required for Masks used as Mask R-CNN input. The reason why the 3M high tack spray adhesive bottle cannot be segment well is that the COCO dataset did not include enough bottle objects. Then we consider adding more bottle objects in the FCN training dataset. After the training dataset of FCN include the some bottle objects, these categories of bottle objects can be segmented correctly, bottle objects segmentations result is shown in Fig 6. Then the object segmentation results are used as the input Masks of Mask R-CNN, the object detection results is shown in Fig 7, (a) show that the blue bottle can be detected with mAP of 0.999,(b)(c)(d) show that the bottle in different orientation was detected with mAP of 0.966,0.958, and 0.826 separately. Further we also verify our proposed method to detect object with different view, in Fig 8, (a) show that the 3M high tack spray adhesive bottle can be detected with mAP of 0.996,(b)(c)(d) show that the bottle in different orientation was detected with mAP of 0.972,0.997, and 0.996 separately.

### 4. Conclusion

In this paper, an automatically generating image masks for Mask R-CNN to object detection is proposed. Through our proposed method, the objects are segmented correctly with FCN, then those objects segmentation results are used as input Masks for Mask R-CNN.
The experimental results show that the proposed method achieves an over 90% mAP for segmentation accuracy. The proposed method can gain the masks automatically with FCN, and then those masks are used to train Mask R-CNN. The proposed object detection system is expected to be used in the automatic object detection industry for analysis of defects with detail position and type information.

### 5. Acknowledgment

This research was supported in part by the China National Key Research and Development project (2017YFE0113200), National Natural Science Foundation of China ( No.51605004) ,Anhui Provincial Natural Science Foundation(1808085QE162), Open Project of Anhui Province Key Laboratory of Special and Heavy Load Robot.

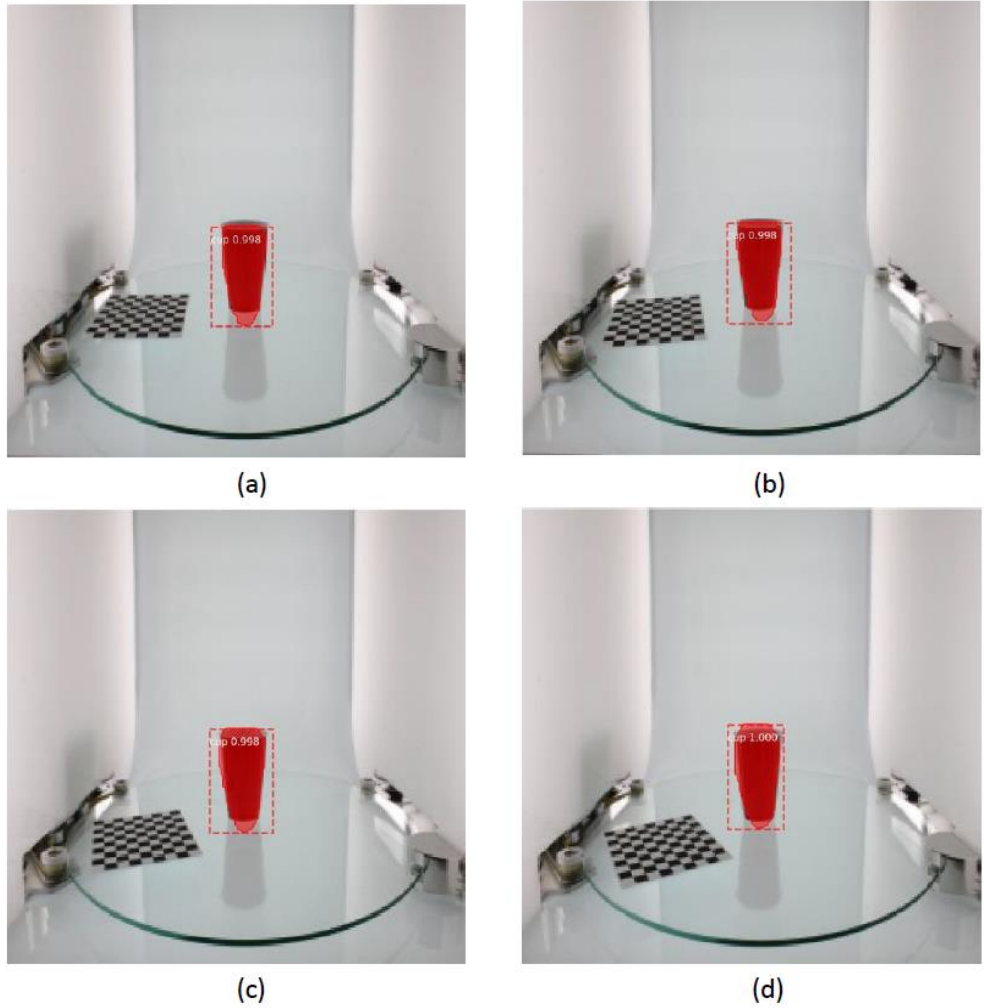

Figure 4. Detection results for cup with Mask R-CNN

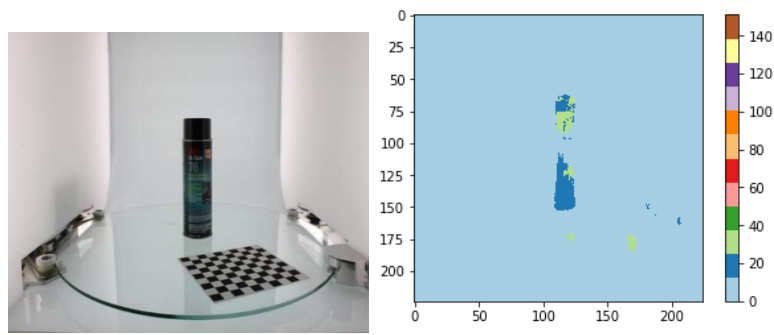

Figure 5. 3M high tack spray adhesive bottle segmentation results

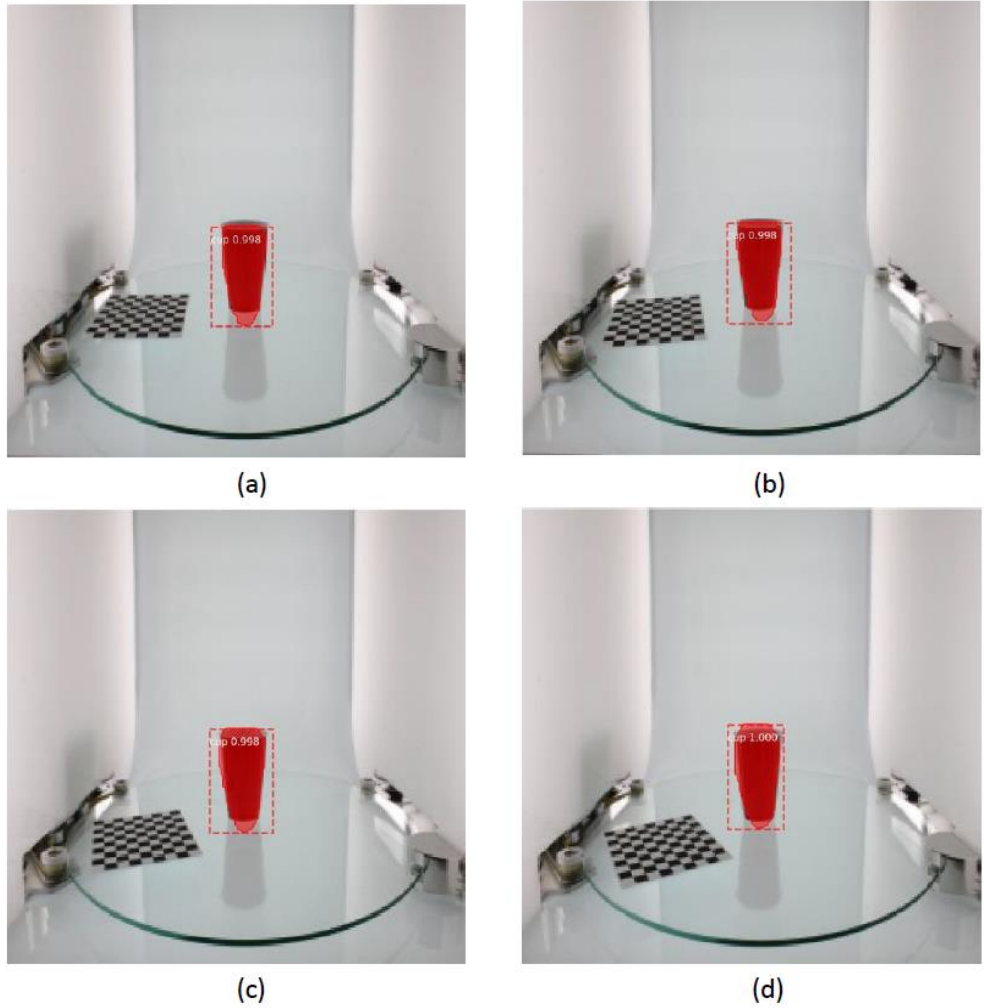

Figure 4. Detection results for cup with Mask R-CNN

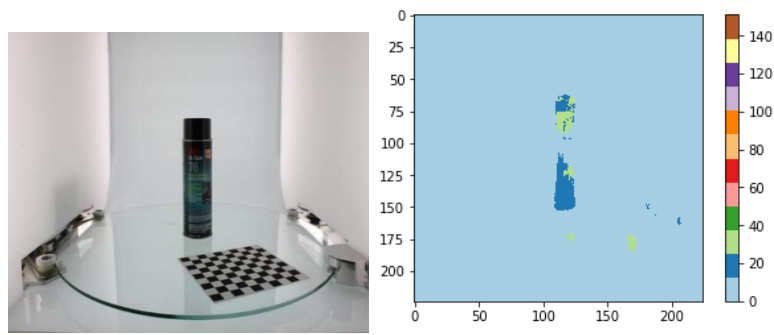

Figure 5. 3M high tack spray adhesive bottle segmentation results

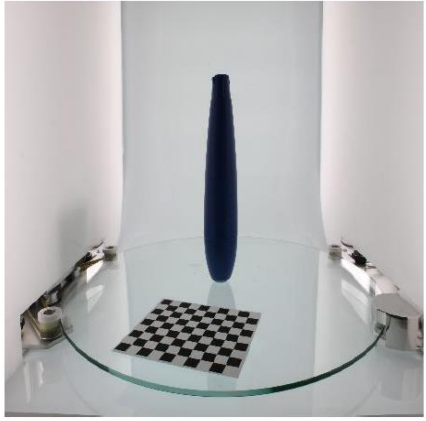
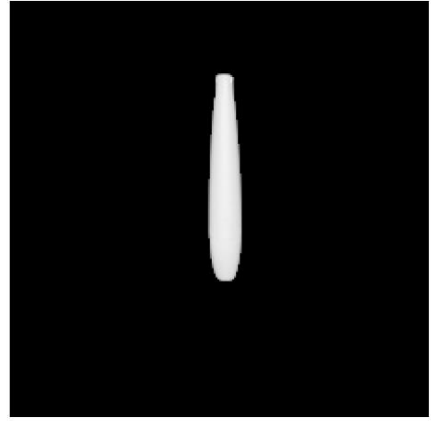

(a)                  (b)

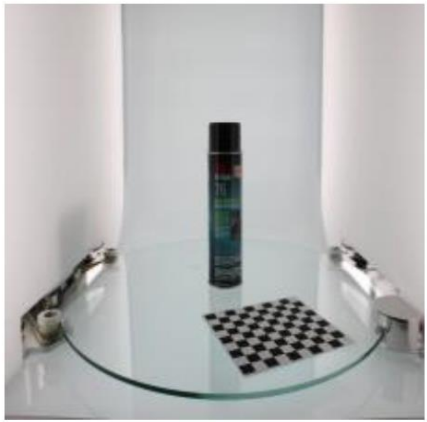
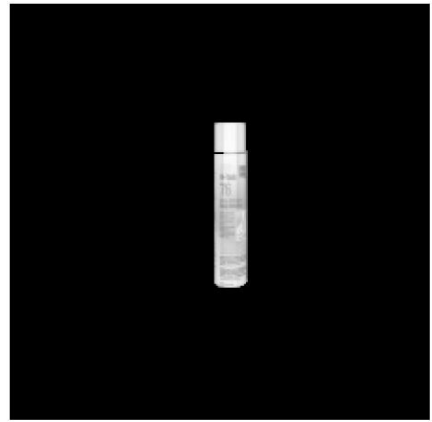

(c)                  (d)

Figure 6. bottle objects segmentations results

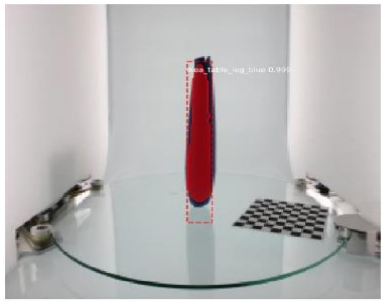
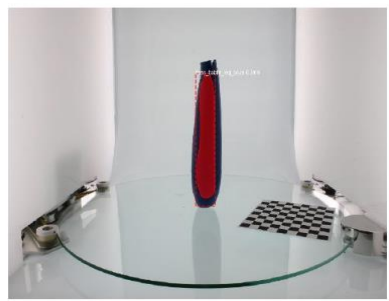

(a)                  (b)

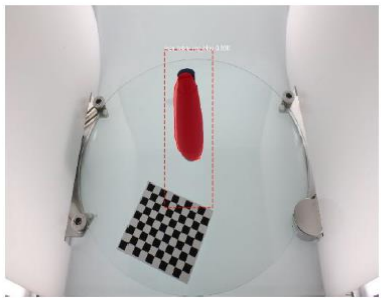
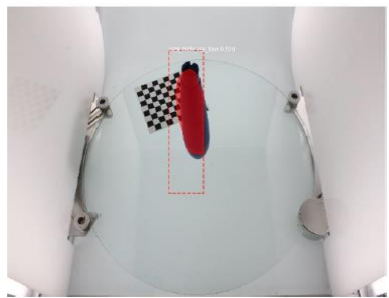

(c)                  (d)

Figure 7. Detection results for blue bottle with different orientation

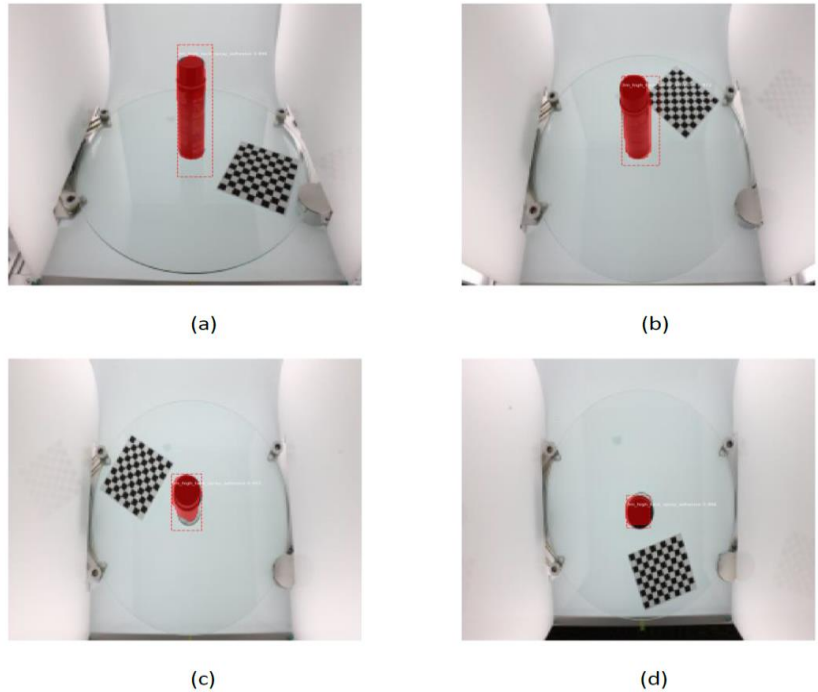

Figure 8. Detection results for 3M spray adhesive bottle with different view